\documentclass[a4paper,fleqn]{cas-dc}

\usepackage[numbers]{natbib}
\AtBeginDocument{\AtBeginShipoutNext{\AtBeginShipoutDiscard}}
\def\tsc#1{\csdef{#1}{\textsc{\lowercase{#1}}\xspace}}
\tsc{WGM}
\tsc{QE}
\tsc{EP}
\tsc{PMS}
\tsc{BEC}
\tsc{DE}

\begin{document}
\let\WriteBookmarks\relax
\def\floatpagepagefraction{1}
\def\textpagefraction{.001}
\shorttitle{Probabilistic analysis of solar cell optical performance using gaussian processes}
\shortauthors{Jaiswal et~al.}

\title [mode = title]{Probabilistic analysis of solar cell optical performance using gaussian processes}                      



\author[1,2]{Rahul Jaiswal}

\address[1]{Center for High Technology Materials, Albuquerque}
 
\author[2]{Manel Martinez Ramon}
\author[1,2,3]{Tito Busani}

\address[2]{Electrical \& Computer Engineering Department, University of New Mexico}
\address[3]{Corresponding author}

\



\begin{abstract}
This work  investigates application of different machine learning based prediction methodologies to estimate the performance of silicon based textured cells. Concept of confidence bound regions is introduced and advantages of this concept are discussed in detail. Results show that reflection profiles and depth dependent optical generation profiles can be accurately estimated using gaussian processes with exact knowledge of uncertainty in the prediction values.It is also shown that cell design parameters can be estimated for a desired performance metric.

\end{abstract}




\maketitle

\section{Introduction}

Machine learning based methods are being used to optimize photovoltaic device design and fabrication recipes \cite{wagner-mohnsen-2020}, which has shown to be better in terms of resource, manpower and time expenditure than traditional loss optimization strategies which involve repetitions of experiments to yield quantitatively varying data-sets. Although, there are factors \cite{LIU2017159} that restrict the widespread acceptance of using machine learning model predictions for device and material optimization like their accuracy, which is generally estimated after  model training on a test database, and generalizability \cite{chung2019unknown}, which is unknown until compared to characterization/simulation data.

Gaussian process for machine learning are used in tthis research \cite{rasmussen_i._2008}. In particular, we apply Gaussian process regression (GPR). The GPR methodology is based a Gaussian model for the training data error , which leads to a Gaussian likelihood function for the regressors and, together with a Gaussian assumption for the prior distribution of the model parameters, it leads to a model for the prediction that includes a mean and a variance of the prediction. This is advantageous over other regression approaches because provided the likelihood and the prior assumptions are correct, they provide not only a prediction but also a confidence interval over this prediction that allows to determine whether the quality of this prediction is acceptable or not for the application at hand \cite{861310}.   Also, the nature of GPR is such that it does not have free parameters, so no cross validation is needed in the training process. 

Using GPR's ensures some degree of confidence that any conclusions made during predictions are robust to the extent of uncertainty in the data, i.e. it makes sure that the inferred parameter (cell performance parameter like reflection profile or cell design parameter like texture angle) is not specific to the particular noisy dataset that was used for training the model, and a confidence bound for a given number of standard deviations can be drawn over the regression predictions. These confidence bounds for each prediction can be monitored to dynamically inform decisions about when to trust a trained model’s predictions in a high throughput environment. In our work GPR prediction mean values are also compared against the prediction of other regression models like ensemble techniques and neural networks.


Complete numerical simulation analysis for a solar cell can be divided into optical and electrical simulation, as the unit cell for a textured cell in electrical domain will contain a large number of repeating pyramid structures. In this article, machine learning model equivalents for optical simulations are developed and uncertainty of model prediction is studied for different test cases (Different cell design and material parameters). Model generalization for other training data sets (Experimental data) is discussed. Apart from forward prediction (i.e., outcome of a process step, given the process parameters and the results of the previous process step), the efficiency of back prediction (predicting the change in design/material parameters to achieve a desired process result) is also explored, along with its uncertainty. Finally , a strategy to deploy trained machine learning models is discussed.


\section{Methodology}
For calculating the optical performance of solar cells, the software Sentaurus device TCAD is used , python was used for machine learning model development and statistical analysis. This work involves three main steps. Initially, a set of variables (Cell design and material parameters) for optical simulation are used as input for the Sentaurus model to simulate the solar cell optical performance (characterized by optical generation in the cell and reflactance from the cell). Using this data set (Simulation parameters and corresponding outputs), machine learning models are trained. Finally,  these models are used to predict cell performance for certain test cases, and apart from prediction accuracy, the uncertainty of prediction is observed for test cases that are withing the training data set range and outside of it. Confidence interval in these predictions are also quantified for input test points that are close to training points and for those which are comparatively far away. 
\subsection{Device Model}

\begin{figure}[ht]
    \centering
    \includegraphics[width=.45\textwidth]{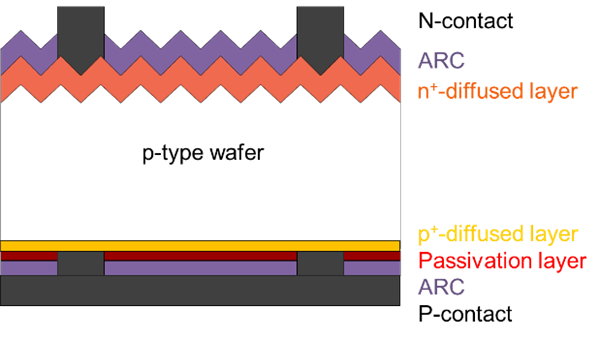}
    \centering
    \caption{\centering PERC cell}
    \label{fig:mesh1}
\end{figure}

p-PERC cell with totally diffused rear \cite{8547393} (figure 1) was the target design that was chosen in this study. A three dimensional Sentaurus TCAD model was designed for optical simulation on a single side textured structure (with transfer matrix method applied for the thin front nitride layer and ray tracing for rest of the wafer) and free carrier absorption \cite{fca} was enabled. Among all the parameters that effect the optical structure, we choose the following six parameters, with known physical effects on the cell optical performance to be varied in the simulation for creating a simulation database:

\subsubsection{Wafer Thickness}
We are simulating two performance metrics: reflection profile and depth dependent carrier generation, which accommodates effects of incident light rays of shorter  and longer wavelength absorption for different cell thicknesses\cite{ZAKI201771}. The optical generation and the reflection from the cell are both dependent on the path length that the incident light travel within the wafer.
\subsubsection{Substrate doping}
Free carrier absorption within the wafer has a dependency on carrier concentration and is observed only for high doping concentrations, we have included this parameter in our study for two reasons, first to see machine learning predictions agree with knowledge from literature and simulation results and second, the result of this optical simulation prediction/simulation will be fed into a electrical simulation/prediction model, so compensating effect of a parameter change in two prediction models can be studied in future.
\subsubsection{Front textured pyramid angle}
The simulation unit cell was composed of upright regular pyramids, this unit cell had perfect reflecting boundaries (To virtualize the full cell structure). Pyramid structures helps in incident light trapping via multiple reflections and total internal reflection \cite{6186552}. Incident light scattering will be different for different pyramid dimensions, dependent on the pyramid base angle.
\subsubsection{Rear side contact (aluminum) thickness}
The aluminum-silicon interface at the rear of Si solar cells absorbs long wavelength photons not collected in the Si photovoltaic (PV) absorber, eliminating the transmittivity through the cell. Variation in the thickness of this contact should not have any effect on cell optical performance. This concept will be studied by comparing the machine learning predictions to simulation results during testing. Long wavelength photons may
be converted into heat and reduce operating device efficiency, this concept can be tested in future work using a multi-physics simulation model.
\subsubsection{Front side anti-reflective coating thickness}
The amount of reflectivity from an anti-reflective material coated surface is dependent on thickness of the ARC (anti-reflection coating), which in turn is dependent on the wavelength of the incoming wave and refractive indices of the materials involved. Variation in this thickness can help to identify the optimized ARC thickness \cite{DUTTAGUPTA201278} for a particular cell design rapidly using machine learning models for a polychromatic light source.
\subsubsection{Back reflectivity}
The rear side dielectric layer in PERC cells not just passivates the surface and improve electrical performance, but it reflects the photons that have not been involved in electron hole generation, essentially giving them a second chance to increase the optical generation in the cell. It also contributes to further randomising light (Within the cell). After identifying the optimal dielectric layer material and thickness for rear side, its effect (lumped by the back reflectivity parameter) on improving the cell optical performance \cite{HOLMAN2014426} can be prototyped rapidly using a machine learning model.
\subsection{Data Preparation}
These six input parameters of the simulation models were statistically varied to create 768 simulations. Simulation output for the depth dependent carrier generation profile is a two dimensional list, where the first column contains depth points within the wafer and the second column contains the carrier generation at the corresponding depth points. The value of the last row in the first column (depth points) is equal to the wafer thickness (i.e. one of the input parameters for the simulation). One simulation corresponds to a input matrix of rank $[m \times n]$, where m is 6 (i.e. the number of simulation inputs) and n is 1 (one simulation), while the output matrix has a rank of [r x s], where r is 2 (i.e. two columns, depth and carrier generation) and s is a number distributed between 0 to the wafer thickness value.

The machine learning regression models we were designing predicts one parameter, so we needed to flatten the output list of the simulations to create a training database. The depth points from the simulation output were used as one of the inputs in the training data base (apart from the existing six input parameters). Therefore, one simulation of input rank [m x n] and output rank [r x s] correlates to input matrix of [(m+1) x s] and output matrix of [1 x s] in the training database. The input parameters in one simulation are padded 's' times and a new column (Depth points) are added to create the training database input, while the output matrix in the training database is just a one dimensional list of carrier generation. 

Simulation output for the reflection profile is also a two dimensional list, where the first column contains wavelength of light incident on the wafer and the second column contains the percentage of light rays reflected corresponding to that wavelength. One simulation corresponds to a input matrix of rank [a x b], where a is 6 (i.e. the number of simulation inputs) and b is 1 (one simulation), while the output matrix has a rank of [o x p], where o is 2 (i.e. two columns, wavelength and reflectance) and p is a 18 (The wavelength values are varied from 300 nm to 1.2 um in steps of 50 nm). The output list of the simulations was flattened to create a training database. The wavelength points from the simulation output were used as one of the inputs in the training data base . Therefore, one simulation of input rank [a x b] and output rank [o x p] correlates to input matrix of [(a+1) x p] and output matrix of [1 x p] in the training database.

Finally for verification of trained models, test cases were created, where input feature values were slightly deviated from the training data values and prediction from machine learning models were compared against the simulated results.

A total of 13824 training data points for reflactance profile prediction and 172000 data points for optical generation profile prediction were created. These data points were sampled during model training and testing.

\subsection{Gaussian process regression models}

For reflactance profile prediction, the machine learning model will be trained on the training data-set created (as described in the previous section).The reflectance at a given wavelength can be predicted at a time, by repeating the prediction for different wavelengths (keeping the other six parameters same), the complete reflection profile can be predicted.

Similarly for optical generation profile prediction, generation at a given depth point is predicted at a time, and then this process is repeated for different depth points (i.e. the surface or 'O' to substrate thickness).

In order to proceed with the prediction tasks, a Gaussian process (GP) regression model is used. The GP model is an estimator of the form \cite{rasmussen2003gaussian}
\begin{equation}
y_i={\bf w}^{\top}\phi({\bf x}_i) + e_i
\end{equation}
with $1 \leq i \leq N$, where $y_i$ is the target to be predicted or regressor, ${\bf x}_i$ is the input observation or predictor, and $e$ is the prediction error. From a GP standpoint, the error is considered a sequence of independent and identically distributed Gaussian samples of zero mean and variance $\sigma_n^2$. Function $\phi(\cdot)$ maps the input features into a higher dimensional Hilbert space endowed with a dot product $K({\bf x}_i,{\bf x}_j)=\phi^{\top}({\bf x}_i)\phi({\bf x}_j)$. Funcion $K$ is called a kernel, and, by virtue of the Mercer's theorem \cite{mercer1909xvi}, the only condition for it to be a dot product in a higher dimension Hilber space is that the function is definite positive. The Representer Theorem \cite{scholkopf2001generalized} assures that there exists an equivalent representation of the model into a dual space expressed only as a linear combination of dot producs, whit the form
\begin{equation}
    y_i=\sum_{j=1}^N \alpha_j K({\bf x}_i,{\bf x}_j) + e_i
\end{equation}
where ${\bf x}_j$ is a set of training data, and \(\alpha_j \) is a set of dual trainable parameters. A quantity of positive definite functions can be used as kernel, so the estimator has nonlinear capabilities. 

For our work we have used squared exponential kernel \cite{9248120} (Radial basis function) given by 
 
 \begin{equation}
          K{(\mathbf{x_i}, \mathbf{x_j})} = \sigma_f^2 \exp \left( -\frac{1}{2}
          (\mathbf{x_i} - \mathbf{x_j})^\top \Theta^{-2} (\mathbf{x_i} - \mathbf{x_j}) \right)
       \end{equation}
and rational quadratic kernel \cite{6958899} given by
       \begin{equation}
          K{(\mathbf{x_i}, \mathbf{x_i})} =   \sigma_f^2 \left (1 + \frac{1}{2\alpha}
          (\mathbf{x_i} - \mathbf{x_j})^\top \Theta^{-2} (\mathbf{x_j} - \mathbf{x_j}) \right)^{-\alpha}
       \end{equation}
 where $\sigma_f$, $\alpha$ and $\Theta$ (length-scale) are hyper-parameters of the kernel functions.  

The GP is solved by first stating a prior probability distribution $p({\bf w})$ for the primal parameters $\bf w$, which is a multivariate Standard, and a Gaussian conditional likelihood $p(y_j|{\bf x}_j,{\bf w})$ for the training data, with variance $\sigma_n^2$. By using the Bayes rule, a posterior distribution of the primal parameters is found. Then, a posterior distribution can be found for a test sample ${\bf x}^*$, which is another Gaussian with mean and variance given by
\begin{equation}
    \bar{f}({\bf x}^*) = {\bf y}^{\top}\left({\bf K} +\sigma_n^2{\bf I}\right)^{-1}{\bf k}({\bf x}^*)
\end{equation}
and 
\begin{equation}
    \sigma_*^2 = k({\bf x}^*,{\bf x}^*)-{\bf k}^{\top}({\bf x}^*)\left({\bf K} +\sigma_n^2{\bf I}\right)^{-1}{\bf k}({\bf x}^*)
\end{equation}
where $\bf y$ is a column vector containing all training regressors, $\bf K$ is the kernel matrix of dot products betweeen training predictors $K({\bf x}_i,{\bf x}_j)$, ${\bf k}({\bf x}^*)$ is a column vector containing the dot products  $K({\bf x}^*,{\bf x}_j)$ between the training data and the test sample, and $\bf I$ is an identity matrix. The variance gives a confidence interval over the prediction. The hyperparameters of the kernel and the noise parameter $\sigma_n^2$ are optimizing by maximizing the marginal log likelihood of the training regressors with respect to them, which is usually done by gradient descent.

\section{Results}

\begin{figure}[ht]
    \centering
    \includegraphics[width=.45\textwidth]{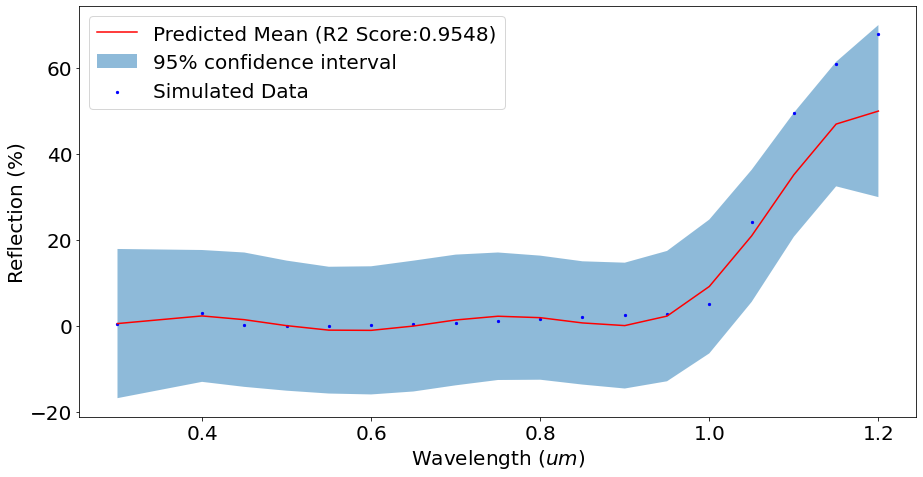}
    \caption{\centering Reflection profile prediction (GP Model trained with 3 features)}
    \label{fig:mesh2}ResultsSquared
\end{figure}

Squared exponential (RBF) kernel was used for the Gaussian process model to predict the reflection profile, With only 3 features used for training the Gaussian process regression model for predicting the reflection profile, the accuracy of mean prediction and confidence bound will be affected, ``Substrate thickness'', ``Substrate doping'' and ``rear side contact (aluminum) thickness'' were the three parameters which were used to train the first model and its prediction for a test case is shown in Figure \ref{fig:mesh2}. 

\begin{figure}[ht]
    \centering
    \includegraphics[width=.45\textwidth]{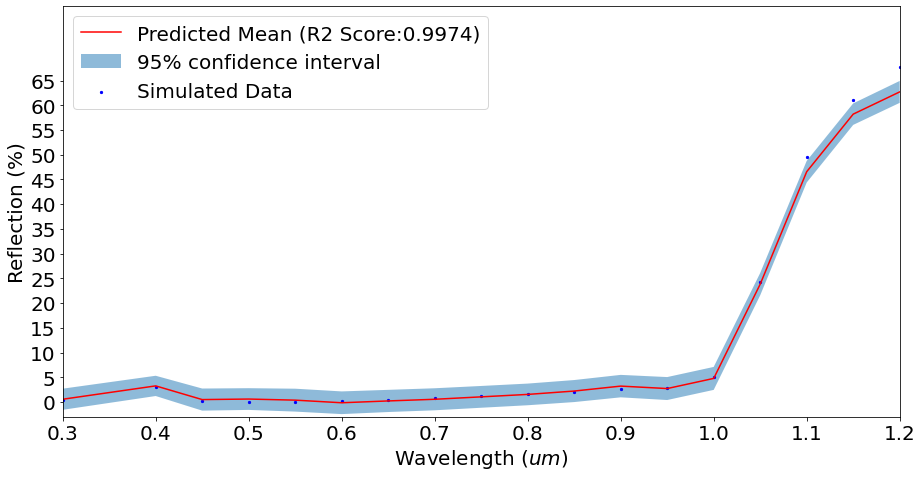}
    \caption{\centering Reflection profile prediction (GP Model trained with all features)}
    \label{fig:mesh3}
\end{figure}

When all six feature values were used for training the prediction accuracy for mean values increased from 0.9596 to 0.9974 and the confidence bound was narrowed as shown in Figure \ref{fig:mesh3}.

Another expected result was obtained when the feature parameter ``rear side contact (aluminum) thicknes''s was excluded from the model training (and the rest of 5 features were used), it has no effect on the model prediction accuracy and confidence interval, which agrees with cell device physics as variation in this parameter should not have any effect on the reflection profile of the wafer.

These same phenomena are observed from machine learning models for optical generation profile prediction, for this model, a rational quadratic (RQ) kernel was used to calculate the covariance, with just 3 input features the prediction accuracy was smaller than that observed for a model trained with all 6 features, the confidence interval was also bigger for the model with 3 input features, these two predictions for a test case is shown in figure 4. Similar to the prediction of reflection profile exclusion of the "``rear side contact (aluminum) thickness'', the feature value has no effect on the prediction accuracy of optical generation profile.

\begin{figure}[ht]
    \centering
    \includegraphics[width=.45\textwidth]{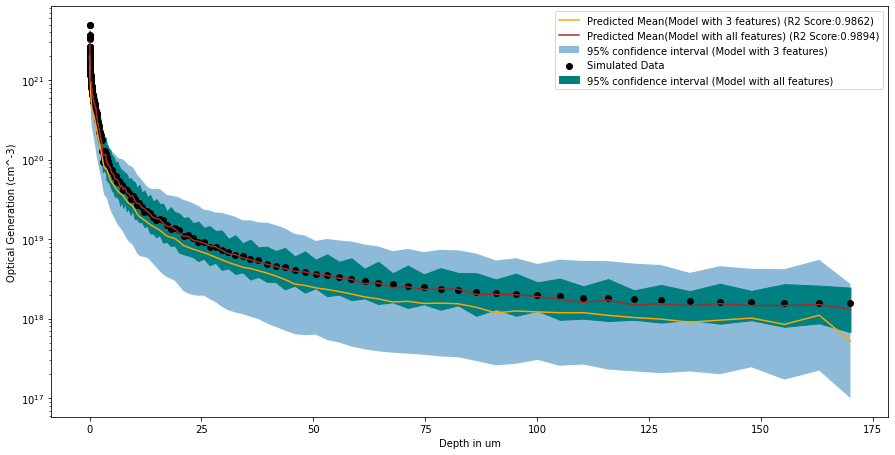}
    \caption{\centering Optical generation profile prediction comparison (GP Model trained with all features vs model trained with 3 features)}
    \label{fig:mesh4}
\end{figure}

\begin{figure}[ht]
    \centering
    \includegraphics[width=.45\textwidth]{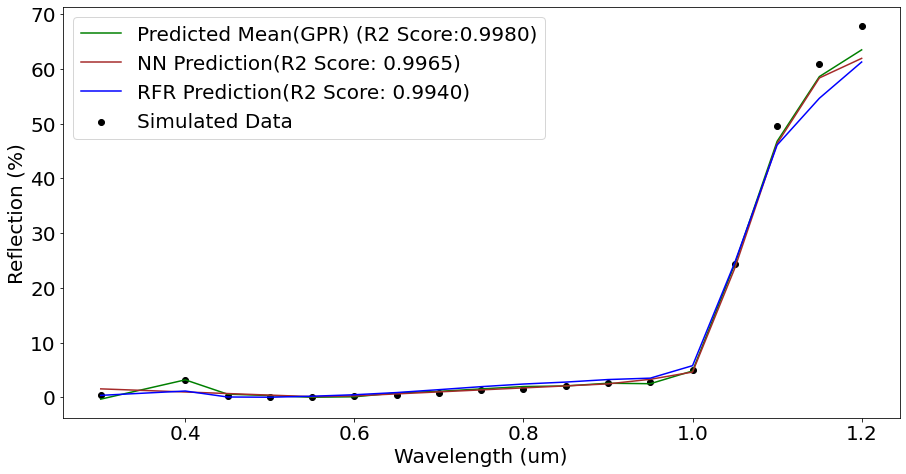}
    \caption{\centering Gaussian process, neural network and random Forest regression prediction for reflactance profile}
    \label{fig:mesh5}
\end{figure}

The mean prediction fromGPR (which is a kernel regression) was also compared to  prediction of other machine learning models. We designed an ensemble model using random forest regression \cite{schonlau-2020} and a neural network regression model \cite{gulli_pal_2017} (3 layers, first two layers have rectified linear activations,  and a linear last layer). This comparison for reflectance profile prediction is shown in Figure \ref{fig:mesh5}.

\begin{figure}[ht]
    \centering
    \includegraphics[width=.45\textwidth]{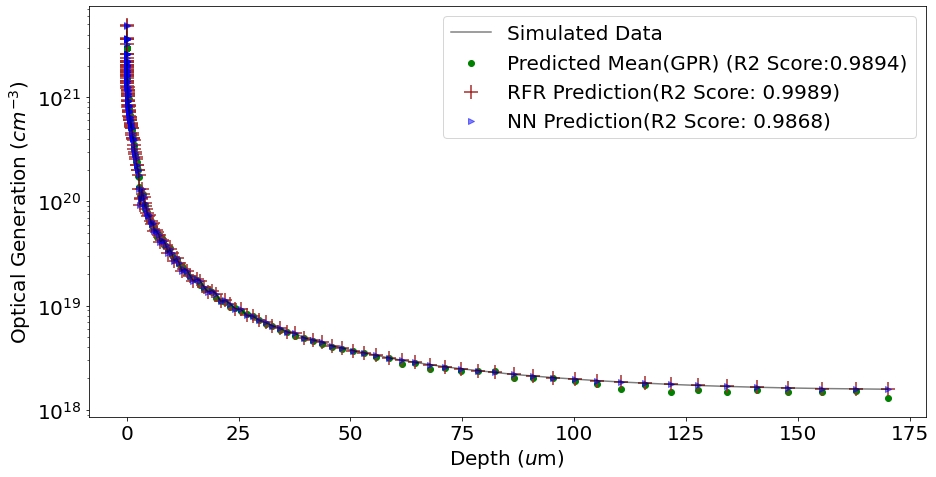}
    \caption{\centering Gaussian process, neural network and random Forest regression prediction for optical generation profile}
    \label{fig:mesh6}
\end{figure}

The comparison of Gaussian process prediction for optical generation profile with neural network and random forest regression prediction is shown in Figure \ref{fig:mesh6}.\\ 

To test the performance of the GPR for predicting other cell performance parameters, we designed two-dimensional electrical device simulation models in Sentaurus TCAD to calculate the current-voltage and minority carrier lifetime profiles. The following parameters were statistically varied in these simulation models to create training databases: substrate doping, substrate thickness, carrier recombination velocities (in the front and rear surfaces), electron and hole lifetime in the wafer, peak rear diffusion and front texture quality (pyramid angle).

\begin{figure}[ht]
    \centering
    \includegraphics[width=.45\textwidth]{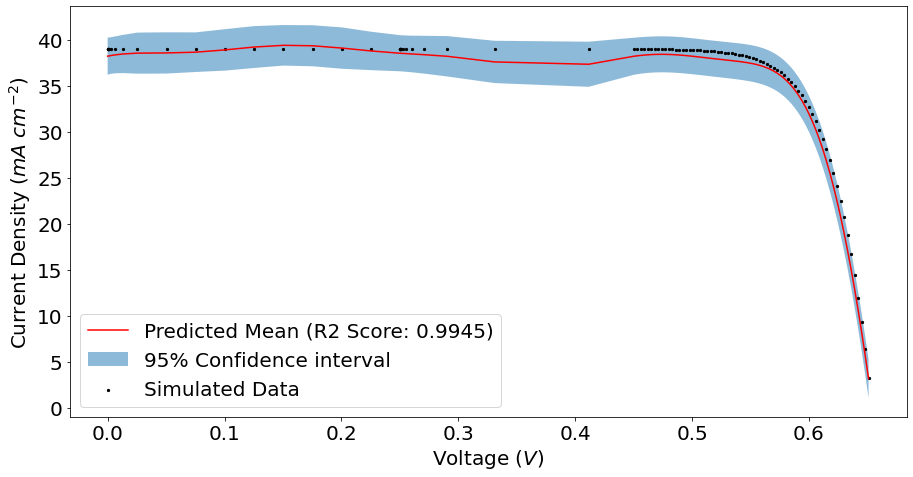}
    \caption{\centering Prediction (Mean value and confidence interval) of cell current voltage profile}
    \label{fig:mesh8}
\end{figure}

\begin{figure}[ht]
    \centering
    \includegraphics[width=.45\textwidth]{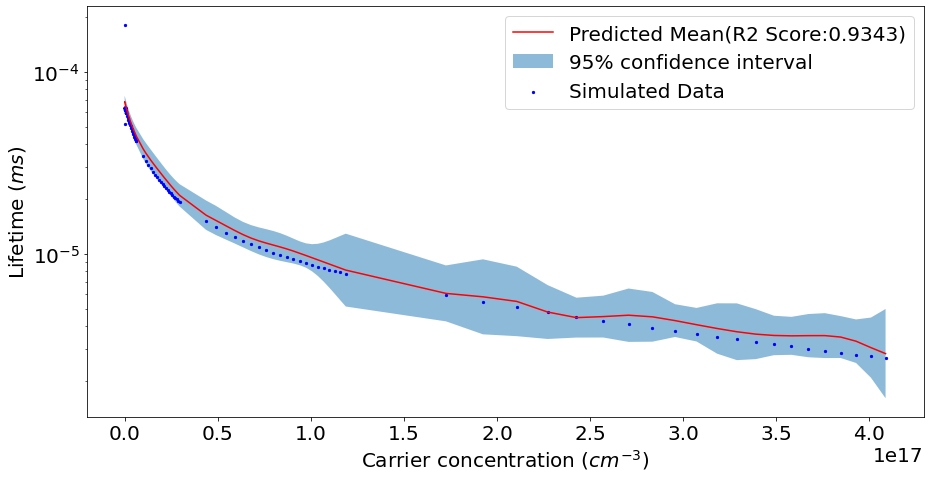}
    \caption{\centering Prediction (Mean value and confidence interval) of minority carrier lifetime in the cell}
    \label{fig:mesh9}
\end{figure}

GPR predictions of cell current voltage profile and minority carrier lifetime profile for a test case are shown in Figures \ref{fig:mesh8} and \ref{fig:mesh9}, and these predictions are compared with their corresponding actual (simulated) values.

Gaussian process regression models can also perform back-prediction(to estimate cell design parameter for a desired performance metric) it can predict both the mean value and confidence interval (uncertainty in back prediction). We trained a model with 6 input parameters - pyramid angle, front side ARC Thickness,back reflectivity, rear side contact thickness,incident light wavelength and the reflection from the cell at that wavelength. The output of the model was cell thickness. Basically, for a given wavelength, we can estimate the wafer thickness necessary to achieve a certain reflection (Given other 5 parameters are kept constant for a single prediction). Figure \ref{fig:mesh9} demonstrates three such predictions for three different reflection values.

\begin{figure}[ht]
    \centering
    \includegraphics[width=.45\textwidth]{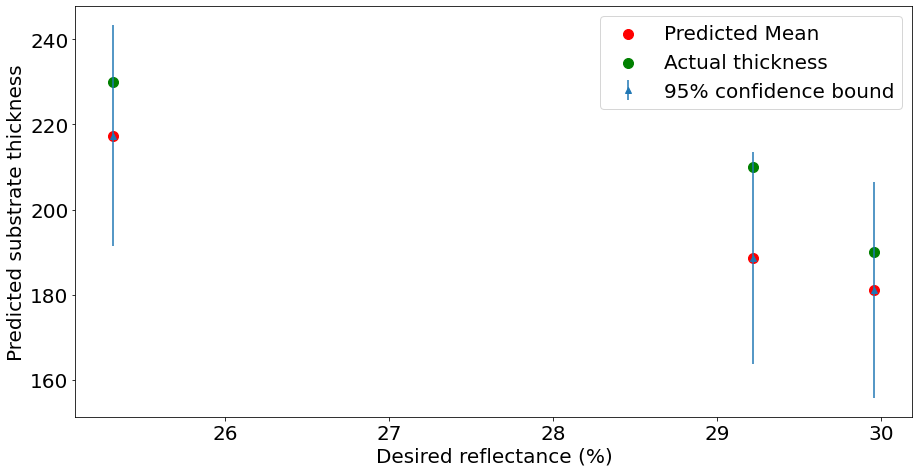}
    \caption{\centering Back predicting wafer thickness necessary to achieve a certain reflection (At a given wavelength and other cell design constraint)}
    \label{fig:mesh10}
\end{figure}

Once trained, the state (hyperparameters in case of gaussian process regression) of the machine learning models can be saved as it is in a static external file (for example, a HDF5 file format). This model can then be wrapped as a REST API (we used the Python Flask framework), and can be made accessible via a web-interface \cite{singh-2021} (for example, a website with form based inputs, where user can enter the model features/cell design parameters) and the model will predict outputs on the fly by loading the pre-calculated model state. The Output (prediction results) can be displayed either as raw data (for example a JSON object) or a plot (image), the framework is described in Figure \ref{fig:mesh8}. This deployment strategy has two main advantages. First, computation is done on a network server (no software setup on the user-side), and second, the model source code and training data is not exposed. We have designed one such API, currently running on our internal website.

\begin{figure}[ht]
    \centering
    \includegraphics[width=.45\textwidth]{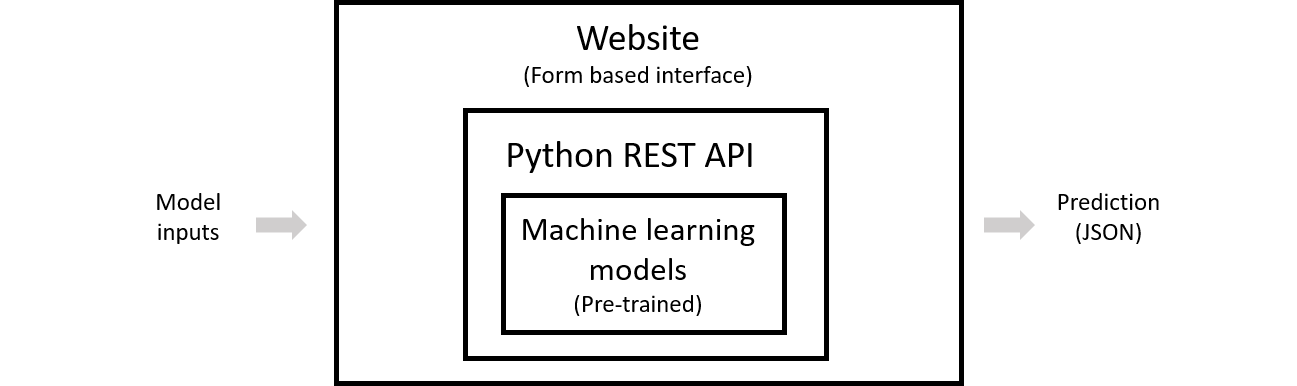}
    \caption{\centering Proposed framework for deployment of trained machine learning model}
    \label{fig:mesh11}
\end{figure}

\section{Discussion}

Machine learning model frameworks proposed in literature \cite{8980687} generally provide a conditional mean value as their prediction. In addition to the mean prediction value, GPR is also providing a confidence interval for cell performance parameters (Optical generation, reflection , minority carrier lifetime and current voltage profile). Other machine learning models like neural networks or random forest regression can also predict the mean values with high accuracy, but there is no covariance calculation in these models.

GPR prediction for models trained with different input feature confirms that features with known effect on cell performance from literature will significantly affect the prediction quality. These model will have a better accuracy to predict the mean value and is more confident about it (i.e. the confidence interval width is narrower) when these input features are included in the training.  Also, exclusion of features which don't have any effect on cell optical performance for model training does not affect the quality of prediction. This conclusion is indicating that one can increase the number of input features (More than what is used in our study) for model training to potentially improve the prediction accuracy and confidence intervals.

We found in our analysis that GPR can also be used to predict cell electrical performance parameters, and an important feature of GPR is observed in the prediction of current-voltage and lifetime profile, when the model is trained on data which is sparsely distributed between a range of input features values, the model will be less confident to predict for input parameters within this range, i.e. the prediction mean will start approaching the prior mean (set during model definition) and the confidence interval will be large. This effect is observed in current voltage profile prediction as the simulation model calculates current-voltage values sparsely before the maximum power point (as there is not much variance in this region) and around the maximum power point the calculation density is highest (to accurately predict the maximum power point). Therefore the training data produced has least number of data points in the 0.2 to 0.45 volts region, and consecutively the machine learning model trained on this data is comparatively less confident (Wider confidence interval) in predicting current values in this range, the accuracy for mean value prediction is also less(compared to its accuracy in other input parameter regions). This phenomenon is even more exaggerated in the model for predicting minority carrier lifetime, as there is no training data points around carrier concentration of \(1.5 * 10^{17}\) \(cm^-3\), so the prediction confidence interval is very wide around this point.

\bibliographystyle{unsrtnat}

\section{Conclusion and future work}
GPR can predict solar cell electrical and optical parameters over a wide range of input parameters and the prediction quality depends on training data and input features, so hypothetically these models can be retrained using experimental (measured data) or a combination of simulation and characterization data for a given cell architecture. Given a performance metric value, it is also possible to back-predict cell design parameter with uncertainty in the prediction, these features of GPR models make them an ideal prediction tool in an academic or industrial setup where variation in performance of a particular cell architecture, by changing its design parameters can be predicted within a finite range (confidence bound). 

We are also designing 3-dimensional electrical simulation model in Sentaurus TCAD and their digital twins(equivalent machine learning models), in future we plan to present a pipelined structure where optical performance prediction (Optical generation) from one machine learning model will be fed into other machine learning models predicting electrical performance, thereby investigating compensating effect of cell design parameter in electrical and optical domain.

\bibliography{cas-refs}

\end{document}